\documentclass[runningheads]{llncs}
\usepackage[T1]{fontenc}
\usepackage{graphicx}
\usepackage{makecell}
\usepackage{multirow}
\usepackage{multicol}
\usepackage{booktabs}
\usepackage{amsmath,amssymb,amsfonts}

\usepackage{xcolor}  
\definecolor{myRed}{RGB}{195,10,10}
\definecolor{myGreen}{RGB}{55,149,73}
\definecolor{myBlue}{RGB}{5,5,60}

\usepackage{marvosym} 

\begin{document}
\title{
Training LLMs with Reinforcement Learning over Digital Twin 
Representations for Reasoning-Intensive Surgical VideoQA
}
\titlerunning{Reasoning-Intensive Surgical VideoQA}
%
\author{Yiqing Shen \and Han Zhang \and Mathias Unberath\textsuperscript{(\Letter)}}
%
\authorrunning{Y. Shen et al.}
%
\institute{Johns Hopkins University, Baltimore, MD, USA\\
\email{\{yshen92,unberath\}@jhu.edu}}
\maketitle              

\begin{abstract}
Surgical video question answering requires multi-step reasoning across semantic, spatial, and temporal dimensions.
Existing methods architecturally compress videos into discrete token representations and couple visual perception with reasoning. This approach fragments continuous spatial-temporal relationships and has been shown to restrict multi-step reasoning capabilities.
We introduce a reinforcement learning (RL) framework that trains large language models (LLMs) to decouple perception from reasoning by operating over digital twin representations constructed from surgical foundation models.
Additionally, we introduce hierarchical representations across frame, temporal window, and procedure levels with probabilistic uncertainty estimates.
Finally, we propose a novel reward that combines format validation with accuracy assessment through clinical plausibility evaluation and uncertainty-aware calibration for training.
To demonstrate the capabilities of this approach, we introduce REAL-Colon-Reason, a colonoscopic benchmark with 2000 question-answer pairs across three complexity levels.
We achieve state-of-the-art performance on REAL-Colon-Reason and two existing surgical VideoQA benchmarks REAL-Colon-VQA and EndoVis18-VQA.

\keywords{Surgical Video Question Answering \and Digital Twin Representation \and Reinforcement Learning \and Large Language Model}
\end{abstract}

\section{Introduction}
Surgical video question answering (VideoQA) interprets natural language questions about videos to support automated workflow analysis, trainee education, and many other applications in surgery~\cite{ding2025visual,surgvivqa,surgicalvqla,chen2024llm}.
However, most existing formulations treat this as a direct visual matching task, where models learn to map explicit questions about visible elements to corresponding visual features~\cite{surgicalvqa,liang2024neural}.
Recent datasets such as REAL-Colon-VQA~\cite{surgvivqa} have begun incorporating temporal questions that span multiple frames, yet these still rely on single-step visual recognition rather than multi-step reasoning.
In other words, existing surgical VideoQA neglects scenarios where answering requires multi-step semantic reasoning about instrument function, spatial reasoning to understand geometric/depth relations, or temporal reasoning to trace motion trajectories~\cite{surgvivqa,jit,wang2024grounded}.
The absence of reasoning capacity prevents VideoQA from performing complex surgical tasks, such as anticipating adverse events during safety monitoring or providing guidance when surgeons face ambiguous intraoperative situations~\cite{eppler2023automated}.

On the other hand, existing surgical VideoQA methods do not address these scenarios requiring reasoning~\cite{song2018explore}.
General vision-language models (VLMs) such as Qwen-VL~\cite{qwen2.5vl,qwen3vl} require costly domain-specific adaptation to capture surgical-specific patterns because their visual encoders are trained on natural images.
Most surgery-specific approaches including SSG-VQA~\cite{ssgvqa}, SurgicalGPT~\cite{surgicalgpt} and PitVQA~\cite{pitvqa} operate on single frames, making them unable to track temporal dependencies.
More recent surgical VideoQA methods like SurgViVQA~\cite{surgvivqa} introduce temporal modeling, yet compress visual information into discrete token representations that fragment continuous spatial-temporal relationships, thereby restricting their ability to perform multi-step spatial and temporal inference.
Furthermore, all these approaches architecturally bind visual perception to reasoning, forcing the same model to simultaneously learn fine-grained visual feature extraction and high-level reasoning during training~\cite{jit,dtr1}.
It can consequently introduce competing optimization objectives that require large-scale annotated surgical datasets~\cite{dtr1,zhou2025proreason}.

To overcome these limitations, we introduce a reinforcement learning (RL) framework that trains large language models (LLMs) to perform reasoning over structured intermediate representations rather than directly processing video.
Formally, the LLM learns to plan the construction of intermediate representation and subsequently reasons over it via a structured rollout.
We follow previous work~\cite{jit,dtr1,shen2025text,shen2025reasoning} by terming this representation as digital twin (DT) representation because it functions as a virtual replica of the surgical video~\cite{shen2025position}, which is constructed by calling surgical foundation models to extract semantic entities together with their spatial and temporal relations.
Because surgical procedures involve temporal hierarchies ranging from frame-level motions to workflow phases~\cite{yue2023cascade}, our DT representation encodes observations at multiple time-scales through hierarchical structuring.
Additionally, to handle perceptual ambiguity inherent in surgical videos~\cite{liu2025sam2s}, we formulate entities and relations in the DT representation as probabilistic instead of deterministic, carrying uncertainty estimates.
Finally, we design rewards that enforce clinical validity and calibrate answer confidence to DT uncertainty, eliminating the need for manually annotated reasoning chains.

Our major contributions are four-fold.
First, we introduce an RL framework that trains LLM to decouple perception from reasoning by planning and operating over DT representations constructed by surgical foundation models.
Second, we design hierarchical and probabilistic DT representations that encode multi-scale temporal structures and uncertainty estimates to address the temporal hierarchies and perceptual ambiguities in surgical video analysis.
Third, we develop a reward that enforces clinical validity and calibrates answer confidence to uncertainty.
Fourth, we introduce REAL-Colon-Reason, which is a colonoscopic benchmark for reasoning-intensive surgical VideoQA. 
It includes questions that require multi-step semantic, spatial, and temporal reasoning across varying complexity levels.

\section{Methods}

\subsubsection{Structure of Rollout Sequence.}
\begin{figure*}[t!]
\centering
\includegraphics[width=0.95\linewidth]{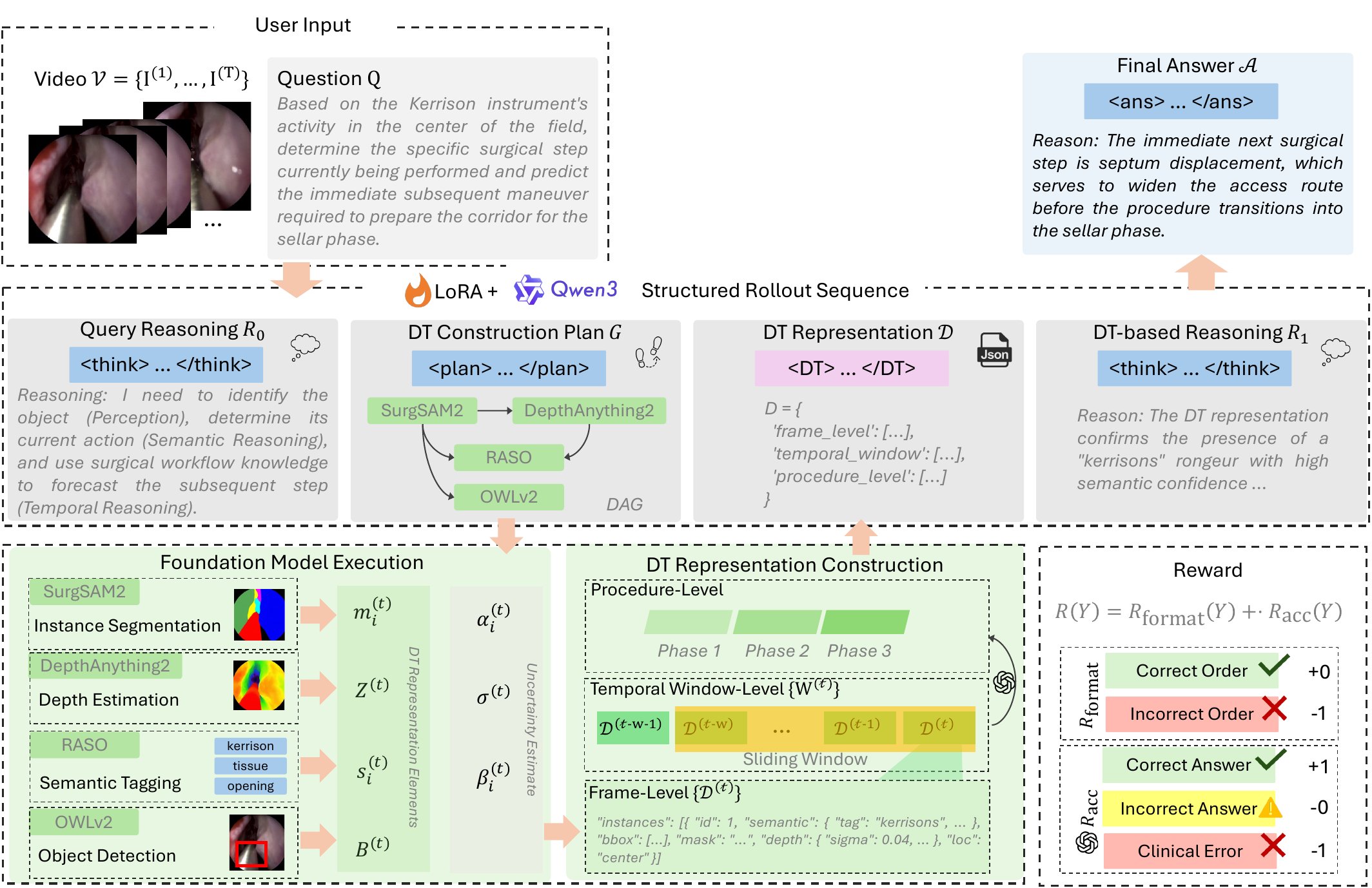} 
\caption{
Overview of the proposed method.
LLM generates structured rollout sequences to plan, construct, and reason over digital twin representations of surgical videos.
}
\label{fig:method}
\end{figure*}

Given a surgical video $\mathcal{V} = \{I^{(1)}, \ldots, I^{(T)}\}$ with $T$ frames and an implicit query $Q$, we train LLM to generate the corresponding answer $A$ via structured rollout sequence, as shown in Fig.~\ref{fig:method}.
The rollout sequence begins with initial query reasoning, where the LLM produces chain-of-thoughts~\cite{cot} $R_0$ between \texttt{$\langle$think$\rangle$} and \texttt{$\langle$/think$\rangle$} tokens to decompose the query and identify required information.
Based on this analysis, LLM then outputs a DT representation construction plan $G$ enclosed by \texttt{$\langle$plan$\rangle$} and \texttt{$\langle$/plan$\rangle$}.
The $G$ specifies which surgical foundation models to invoke and their execution dependencies as a directed acyclic graph (DAG)~\cite{jit,dtr1}, where the LLM receives descriptions of available foundation models within its input prompt.
Once the \texttt{$\langle$/plan$\rangle$} token is detected, LLM generation pause unitl the execution of the designated foundation models.
DT representation $\mathcal{D}$ is then appended between \texttt{$\langle$dt$\rangle$} and \texttt{$\langle$/dt$\rangle$} tokens.
Afterwards, the generation resumes with another reasoning $R_1$ on DT representation within \texttt{$\langle$think$\rangle$} and \texttt{$\langle$/think$\rangle$} tokens.
Finally, the LLM generates answer $A$ enclosed by \texttt{$\langle$ans$\rangle$} and \texttt{$\langle$/ans$\rangle$} tokens.
The complete rollout sequence follows the structure $Y = \langle \langle\text{think}\rangle R_0 \langle/\text{think}\rangle, \langle\text{plan}\rangle G \langle/\text{plan}\rangle, \langle\text{dt}\rangle \mathcal{D} \langle/\text{dt}\rangle$,
$\langle\text{think}\rangle R_1 \langle/\text{think}\rangle$,
$\langle\text{ans}\rangle A \langle/\text{ans}\rangle \rangle$.
%
%

\subsubsection{Hierarchical and Probabilistic Digital Twin Representations.}
The construction of the DT representation is guided by the DAG plan $G = (V, E)$, where the vertices $V$ correspond to surgical foundation models and the directed edges $E$ specify the execution dependencies between them, following the previous work~\cite{dtr1}.
Formally, for each frame $I^{(t)}$ at time $t$ in video $\mathcal{V}$, we first construct frame-level DT representations $D^{(t)}$ by applying the selected models according to the topological order in $G$.
In this work, we consider the following foundation models:
SurgSAM-2~\cite{surgicalsam2} provides instance segmentation masks $M^{(t)} = \{m_i^{(t)}\}_{i=1}^{N^{(t)}}$ where $m_i^{(t)}$ denotes the binary mask for object $i$ and $N^{(t)}$ indicates the number of detected instances; 
DepthAnything2~\cite{depthanything} generates dense depth maps $Z^{(t)}$ capturing spatial relationships; 
RASO~\cite{raso} produces semantic tags $S^{(t)} = \{s_i^{(t)}\}_{i=1}^{N^{(t)}}$ identifying surgical objects; 
OWLv2~\cite{owlv2} outputs bounding boxes $B^{(t)} = \{b_i^{(t)}\}_{i=1}^{N^{(t)}}$ for efficient spatial localization.
To avoid redundant storage of high-dimensional mask arrays within the text-based DT representation, we encode file paths $P^{(t)} = \{p_i^{(t)}\}_{i=1}^{N^{(t)}}$ where each path $p_i^{(t)}$ refers to the location of the corresponding mask $m_i^{(t)}$ on disk.
Importantly, rather than invoking all foundation models uniformly, the LLM selects only those required by the specific query in the plan $G$, reducing computational overhead.
%

Then, we extend the DT representation with probabilistic attributes to quantify the reliability.
This uncertainty information originates from the confidence estimates inherently produced by the foundation models.
Specifically, SurgSAM-2~\cite{surgicalsam2} produces not only binary masks $m_i^{(t)}$ but also predicted IoU scores $\alpha_i^{(t)} \in [0,1]$ that measure segmentation certainty for each instance.
Correspondingly, RASO~\cite{raso} associates each semantic tag $s_i^{(t)}$ with a confidence score $\beta_i^{(t)} \in [0,1]$ derived from the sigmoid probability output of its tag decoder prior to thresholding, reflecting the uncertainty for rare instruments or ambiguous anatomical structures.
When DepthAnything2~\cite{depthanything} is invoked, we also compute the standard deviation to quantify spatial position uncertainty across the masked region.

Finally, surgical procedures can unfold across multiple temporal scales, such as frame-level instrument motions occurring within seconds, tissue interactions span multiple seconds, while workflow phases extend over minutes.
To accommodate this multi-scale temporal structure, we organize the complete DT representation $\mathcal{D}$ hierarchically across three levels.
At the frame level, each $D^{(t)}$ captures instantaneous observations at time $t$.
At the temporal window level, we maintain a sliding context $\mathcal{W}^{(t)} = \{D^{(k)} \mid t - w \leq k \leq t\}$ with window size $w$ that aggregates recent frames to preserve short-term dynamics, including instrument trajectories, tissue deformations, and transient occlusions.
Within this window, object correspondence across frames is established through tracking identifiers $\text{id}_i^{(t)}$ assigned by SurgSAM-2's temporal consistency mechanism~\cite{surgicalsam2}, which links the same entity across consecutive frames based on mask overlap and visual feature similarity.
At the procedure level, we maintain aggregated statistics that summarize long-term context beyond the sliding window by an external LLM (\textit{e}.\textit{g}., GPT), compressing extended temporal sequences into workflow-relevant summaries such as phase transitions and cumulative instrument usage patterns.

\subsubsection{Reward Design.}
To train the LLM with RL, we design a rule-based reward that evaluates both the correctness of rollout structure and the answers.
The total reward $R(Y)$ combines a format reward $R_{\text{format}}(Y)$ and an accuracy reward $R_{\text{acc}}(Y)$, represented as $R(Y) = R_{\text{format}}(Y) +  \cdot R_{\text{acc}}(Y)$.
%
The format reward verifies that the rollout sequence contains all required token pairs in the correct order, which assigns $0$ when all requirements are satisfied and $-1$ otherwise.
The accuracy reward evaluates whether the final answer $A$ within \texttt{$\langle$ans$\rangle$} tokens matches the ground truth reference $A^*$.
Since surgical answers can be expressed in different languages, we employ an LLM-as-a-judge~\cite{llmjudging} approach where a separate evaluation LLM (such as GPT) determines the semantic equivalence between $A$ and $A^*$.
Beyond semantic matching, we instruct the judge LLM to identify clinically implausible claims that violate surgical knowledge, such as contradictory instrument functions, anatomically impossible spatial relationships, or medically unsafe procedural sequences.
Therefore, the accuracy reward is computed as $R_{\text{acc}}(Y) = \mathbb{I}[\text{match}(A, A^*)] - \mathbb{I}[\text{implausible}(A)]$, where the first indicator returns $+1$ when the answers match and $0$ otherwise, while the second indicator imposes a penalty of $-1$ when implausible claims are detected.
This formulation results in final values of $+1$ for correct answers, $0$ for incorrect but plausible answers, and $-1$ for answers containing clinical errors.

To calibrate the confidence of the answer with the uncertainty encoded in the DT representation, we modulate the accuracy reward by aggregating the uncertainty from queries relevant instances.
Let $\mathcal{R} \subset \{1, \ldots, N^{(t)}\}$ denote the set of instance indices to which the LLM refers when generating the answer $A$.
For each instance $i \in \mathcal{R}$, we combine all uncertainty estimates if available: segmentation confidence $\alpha_i^{(t)}$ from SurgSAM-2~\cite{surgicalsam2}, semantic recognition confidence $\beta_i^{(t)}$ from RASO~\cite{raso}, and normalized depth uncertainty $\hat{\sigma}_i^{(t)} = 1 - \sigma_i^{(t)}/\sigma_{\max}$ where higher values indicate greater reliability.
The overall confidence factor is therefore computed as the average reliability across all relevant instances and uncertainty types, denoted as
\begin{equation}
\gamma = \frac{1}{|\mathcal{R}|} \sum_{i \in \mathcal{R}} \left( w_{\alpha} \cdot \alpha_i^{(t)} + w_{\beta} \cdot \beta_i^{(t)} + w_{\sigma} \cdot \hat{\sigma}_i^{(t)} \right),
\end{equation}
where $w_{\alpha}$, $w_{\beta}$, and $w_{\sigma}$ are balancing coefficients, satisfying $w_{\alpha} + w_{\beta} + w_{\sigma} = 1$.
The final modulated reward becomes $\tilde{R}_{\text{acc}}(Y) = \gamma \cdot R_{\text{acc}}(Y)$, which scales the magnitude of the reward according to the reliability of visual observation.
Finally, the LLM is trained using Group Relative Policy Optimization (GRPO)~\cite{grpo} with reward function $R(Y)$.

\subsubsection{Dataset Construction.}
To train and evaluate the proposed method for reasoning-intensive surgical VideoQA, we introduce REAL-Colon-Reason based on the colonoscopic REAL-Colon dataset~\cite{biffi2024real}.
REAL-Colon comprises 60 full-resolution colonoscopy videos, which are annotated at the frame level for endoscope motion trajectories, surgical tool interactions, tissue visibility conditions, and lesion characteristics.
We first segment these videos into three to five minute clips by sampling frames at stride ten from the original 30 FPS footage to balance extended temporal coverage with computational tractability.
To construct reasoning-intensive question-answer pairs, we employ a VLM (\textit{i}.\textit{e}., GPT-5.2) to generate candidate questions along with explicit intermediate reasoning steps required to arrive at each answer by taking both the video clip and the frame-level meta information as input.
All generated pairs undergo manual refinement by clinical experts to verify reasoning chain correctness, eliminate ambiguous formulations, and ensure accuracy.
Questions are organized into three complexity levels based on reasoning chain depth, where level 1 involves basic single-step inference, level 2 requires two intermediate reasoning operations, and level 3 demands chains exceeding three steps of logical deduction.
The final dataset contains 2000 question-answer pairs.

\section{Experiments}

\subsubsection{Implementation Details.}
We employ Qwen3-8B~\cite{qwen3} as the backbone LLM and train it using GRPO with LoRA~\cite{lora} of rank 8 with TRL (v0.26).
The training uses AdamW optimizer with a learning rate of $2 \times 10^{-4}$, a batch size of 8, and a sampling of 4 rollouts per query.
We employ GPT-5-nano to judge the semantic equivalence and clinical plausibility in reward.
For DT construction, we include SurgSAM-2~\cite{surgicalsam2}, DepthAnything2~\cite{depthanything}, RASO~\cite{raso}, and OWLv2~\cite{owlv2}, following their official implementations.
We conduct training on 8 NVIDIA NVIDIA 4090 GPUs of 24Gb memory using DeepSpeed for distributed optimization.

\subsubsection{Dataset and Evaluation Metrics.}
We first evaluate our method on the proposed REAL-Colon-Reason dataset, which is partitioned at the video level with an 80\%:20\% split.
During training, we augment our dataset with 4,450 additional question-answer pairs from REAL-Colon-VQA~\cite{surgvivqa} to improve the model's generalization across diverse query formulations.
For REAL-Colon-Reason, we adopt exact match (EM) as the primary metric, which evaluates whether the generated answer semantically matches the ground truth reference through GPT-5-nano.
We also use SMILE~\cite{smile}, which combines sentence-level semantic understanding with keyword-level semantics and lexical matching, as additional metric.
%
Moreover, we conduct evaluation on REAL-Colon-VQA~\cite{surgvivqa} and EndoVis18-VQA~\cite{pierantozzi2025trust,surgicalvqa} datasets following previous work.
For these benchmarks, we report BLEU-4 for lexical precision, ROUGE-L and METEOR for semantic coherence, and keyword accuracy for clinical term correctness.

\begin{table*}[htbp!]
\centering
\caption{Performance comparison on the REAL-Colon-Reason benchmark.
We report EM and SMILE scores as decimal values [0, 1] (mean $\pm$ std). 
Standard deviations are computed over 3 independent runs with different seeds.
Best results are in \textbf{bold}; second-best results are \underline{underlined}.
}
\label{tab:reasoning_results}
\resizebox{\textwidth}{!}{%
\setlength{\tabcolsep}{4pt}
\begin{tabular}{lcccccccc}
\toprule
 & \multicolumn{2}{c}{\textbf{Level 1}} & \multicolumn{2}{c}{\textbf{Level 2}} & \multicolumn{2}{c}{\textbf{Level 3}} & \multicolumn{2}{c}{\textbf{Overall}} \\
\cmidrule(lr){2-3} \cmidrule(lr){4-5} \cmidrule(lr){6-7} \cmidrule(lr){8-9}
\textbf{Model} & EM & SMILE & EM & SMILE & EM & SMILE & EM & SMILE \\
\midrule
SurgicalGPT~\cite{surgicalgpt}     & 0.224\tiny{$\pm$0.005} & 0.282\tiny{$\pm$0.006} & 0.145\tiny{$\pm$0.004} & 0.193\tiny{$\pm$0.005} & 0.082\tiny{$\pm$0.003} & 0.125\tiny{$\pm$0.004} & 0.150\tiny{$\pm$0.004} & 0.200\tiny{$\pm$0.005} \\
PitVQA~\cite{pitvqa}          & 0.451\tiny{$\pm$0.012} & 0.508\tiny{$\pm$0.010} & 0.286\tiny{$\pm$0.009} & 0.352\tiny{$\pm$0.011} & 0.153\tiny{$\pm$0.006} & 0.221\tiny{$\pm$0.008} & 0.297\tiny{$\pm$0.009} & 0.360\tiny{$\pm$0.010} \\
Qwen2.5-VL-3B~\cite{qwen2.5vl}   & 0.302\tiny{$\pm$0.008} & 0.356\tiny{$\pm$0.009} & 0.224\tiny{$\pm$0.007} & 0.281\tiny{$\pm$0.008} & 0.148\tiny{$\pm$0.005} & 0.205\tiny{$\pm$0.007} & 0.225\tiny{$\pm$0.006} & 0.281\tiny{$\pm$0.008} \\
Qwen3-VL-4B~\cite{qwen3vl}     & 0.355\tiny{$\pm$0.010} & 0.412\tiny{$\pm$0.012} & 0.289\tiny{$\pm$0.009} & 0.345\tiny{$\pm$0.011} & 0.194\tiny{$\pm$0.008} & 0.258\tiny{$\pm$0.009} & 0.279\tiny{$\pm$0.009} & 0.338\tiny{$\pm$0.011} \\
Qwen3-VL-8B~\cite{qwen3vl}     & 0.428\tiny{$\pm$0.011} & 0.485\tiny{$\pm$0.013} & \underline{0.452}\tiny{$\pm$0.012} & \underline{0.511}\tiny{$\pm$0.014} & \underline{0.356}\tiny{$\pm$0.010} & \underline{0.423}\tiny{$\pm$0.012} & \underline{0.412}\tiny{$\pm$0.011} & \underline{0.473}\tiny{$\pm$0.013} \\
MedGemma-4B~\cite{medgemma}     & 0.256\tiny{$\pm$0.007} & 0.314\tiny{$\pm$0.008} & 0.182\tiny{$\pm$0.006} & 0.245\tiny{$\pm$0.007} & 0.105\tiny{$\pm$0.004} & 0.168\tiny{$\pm$0.006} & 0.181\tiny{$\pm$0.006} & 0.242\tiny{$\pm$0.007} \\
InternVL3-1B~\cite{internvl3}    & \underline{0.482}\tiny{$\pm$0.012} & \underline{0.535}\tiny{$\pm$0.014} & 0.354\tiny{$\pm$0.010} & 0.412\tiny{$\pm$0.011} & 0.221\tiny{$\pm$0.008} & 0.295\tiny{$\pm$0.010} & 0.352\tiny{$\pm$0.010} & 0.414\tiny{$\pm$0.012} \\
Surgical-LVLM~\cite{surgicallvlm}   & 0.385\tiny{$\pm$0.010} & 0.441\tiny{$\pm$0.011} & 0.312\tiny{$\pm$0.009} & 0.375\tiny{$\pm$0.010} & 0.208\tiny{$\pm$0.007} & 0.272\tiny{$\pm$0.009} & 0.302\tiny{$\pm$0.009} & 0.363\tiny{$\pm$0.010} \\
VideoLLaMA3-2B~\cite{videollama3}  & 0.185\tiny{$\pm$0.004} & 0.241\tiny{$\pm$0.006} & 0.123\tiny{$\pm$0.003} & 0.178\tiny{$\pm$0.005} & 0.064\tiny{$\pm$0.002} & 0.112\tiny{$\pm$0.004} & 0.124\tiny{$\pm$0.003} & 0.177\tiny{$\pm$0.005} \\
SurgViVQA~\cite{surgvivqa}       & 0.405\tiny{$\pm$0.009} & 0.462\tiny{$\pm$0.011} & 0.264\tiny{$\pm$0.008} & 0.328\tiny{$\pm$0.010} & 0.142\tiny{$\pm$0.005} & 0.205\tiny{$\pm$0.007} & 0.270\tiny{$\pm$0.007} & 0.332\tiny{$\pm$0.009} \\
\textbf{Ours} & \textbf{0.653}\tiny{$\pm$0.013} & \textbf{0.708}\tiny{$\pm$0.015} & \textbf{0.584}\tiny{$\pm$0.012} & \textbf{0.642}\tiny{$\pm$0.014} & \textbf{0.515}\tiny{$\pm$0.011} & \textbf{0.589}\tiny{$\pm$0.013} & \textbf{0.584}\tiny{$\pm$0.012} & \textbf{0.646}\tiny{$\pm$0.014} \\
\bottomrule
\end{tabular}%
}
\end{table*}

\subsubsection{Performance on REAL-Colon-Reason.}
Tab.~\ref{tab:reasoning_results} presents the performance comparison across three levels of reasoning complexity on REAL-Colon-Reason.
Our method achieves the best results with an overall EM of 0.584 and a SMILE score of 0.646, outperforming the second-best method Qwen3-VL-8B~\cite{qwen3vl} by 17.2\% and 17.3\% respectively.
%
%
Specifically, our method shows improvements of 15.9\% at Level 1, 13.2\% at Level 2, and 15.9\% at Level 3 in EM over the strongest baseline, namely Qwen3-VL-8B~\cite{qwen3vl}. 
This demonstrates that DT representations effectively preserve the fine-grained spatial-temporal information required for multi-step reasoning.
Image-based methods such as SurgicalGPT~\cite{surgicalgpt} and PitVQA~\cite{pitvqa} show performance degradation across all levels, illustrating that single-frame approaches fail to capture temporal dependencies needed for temporal reasoning.
SurgViVQA~\cite{surgvivqa} achieves an overall EM of 0.270 compared to our 0.584, despite incorporating temporal modeling through masked video encoding.
This supports our hypothesis that token-based compression fragments continuous spatial-temporal relationships and limits multi-step reasoning capability.
Zero-shot general-purpose VLMs including Qwen3-VL-8B~\cite{qwen3vl} and InternVL3-1B~\cite{internvl3} demonstrate competitive performance at level 1 but show steeper degradation level 2 and 3.
This indicates that without an explicit reasoning structure, even powerful general-purpose VLMs cannot fully handle reasoning-intensive surgical scenarios.

\subsubsection{Performance on Existing VideoQA Benchmarks.}

\begin{table*}[htbp!]
\centering
\caption{Performance comparison on REAL-Colon-VQA and EndoVis18-VQA.
We report BLEU-4 (B-4), ROUGE-L (R-L), METEOR (MET), and Keyword Accuracy (K-ACC) in percentages (\%). 
Best results are in \textbf{bold}; second-best results are \underline{underlined}.}
\label{tab:main_results}
\resizebox{\textwidth}{!}{%
\setlength{\tabcolsep}{3pt}
\begin{tabular}{lcccccccccccccccc}
\toprule
 & \multicolumn{8}{c}{\textbf{REAL-Colon-VQA}~\cite{surgvivqa}} & \multicolumn{8}{c}{\textbf{EndoVis18-VQA}~\cite{surgicalvqa}} \\
\cmidrule(lr){2-9} \cmidrule(lr){10-17}
 & \multicolumn{4}{c}{In-Template} & \multicolumn{4}{c}{Out-of-Template} & \multicolumn{4}{c}{In-Template} & \multicolumn{4}{c}{Out-of-Template~\cite{pierantozzi2025trust}} \\
\cmidrule(lr){2-5} \cmidrule(lr){6-9} \cmidrule(lr){10-13} \cmidrule(lr){14-17}
\textbf{Methods} & B-4 & R-L & MET & K-ACC & B-4 & R-L & MET & K-ACC & B-4 & R-L & MET & K-ACC & B-4 & R-L & MET & K-ACC \\
\midrule
SurgicalGPT~\cite{surgicalgpt}     & 14.93 & 47.85 & 52.36 & 33.33 & 12.35 & 42.87 & 50.49 & 46.67 & 29.55 & 58.60 & 57.99 & 4.00 & 2.52 & 43.97 & 44.99 & 11.85 \\
PitVQA~\cite{pitvqa}          & 64.55 & 78.48 & 79.99 & 54.13 & 23.63 & 50.03 & 53.22 & 42.93 & 81.73 & 86.18 & 83.28 & 40.35 & 17.57 & 46.87 & 45.49 & 9.73 \\
Qwen2.5-VL-3B~\cite{qwen2.5vl}   & 4.53 & 45.70 & 46.83 & 49.33 & 0.50 & 38.80 & 40.77 & 50.27 & 19.04 & 57.60 & 68.60 & 51.22 & 11.92 & 52.77 & 63.72 & 46.90 \\
Qwen3-VL-4B~\cite{qwen3vl}     & 6.80 & 48.20 & 51.50 & 54.00 & 2.10 & 40.50 & 43.80 & 52.40 & 22.50 & 59.30 & 70.20 & 52.80 & 13.20 & 53.50 & 64.80 & 48.10 \\
Qwen3-VL-8B~\cite{qwen3vl}     & 8.90 & 50.10 & 54.70 & 58.20 & 3.50 & 42.80 & 46.50 & \underline{54.10} & 26.30 & 61.40 & 72.50 & \underline{54.60} & 15.80 & \underline{55.20} & \underline{66.90} & \underline{50.30} \\
MedGemma-4B~\cite{medgemma}     & 18.09 & 36.92 & 31.86 & 49.47 & 5.31 & 30.96 & 27.80 & 34.13 & 8.23 & 44.32 & 64.19 & 37.90 & 7.74 & 40.74 & 61.59 & 37.86 \\
InternVL3-1B~\cite{internvl3}    & 7.13 & 43.90 & 54.31 & \underline{68.67} & 3.87 & 32.10 & 44.34 & 53.73 & 5.62 & 51.50 & 55.33 & 32.21 & 0.57 & 38.73 & 38.73 & 21.99 \\
Surgical-LVLM~\cite{surgicallvlm}   & 7.90 & 49.80 & 53.80 & 57.20 & 3.20 & 42.00 & 45.80 & 53.80 & 24.60 & 60.80 & 72.00 & 53.50 & 15.00 & 54.80 & 66.20 & 49.80 \\
VideoLLaMA3-2B~\cite{videollama3}  & 2.74 & 33.01 & 42.22 & 17.60 & 1.78 & 26.17 & 35.15 & 21.47 & 3.28 & 52.01 & 54.66 & 40.90 & 0.57 & 38.73 & 39.51 & 35.89 \\
SurgViVQA~\cite{surgvivqa}       & \underline{71.98} & \underline{82.85} & \underline{84.11} & 63.20 & \underline{31.19} & \underline{53.62} & \underline{54.89} & 47.73 & \underline{84.94} & \underline{89.65} & \underline{86.08} & 48.27 & \underline{24.84} & 50.86 & 50.13 & 9.23 \\
\textbf{Ours} & \textbf{75.42} & \textbf{85.10} & \textbf{86.35} & \textbf{71.05} & \textbf{35.80} & \textbf{57.45} & \textbf{58.90} & \textbf{57.12} & \textbf{87.20} & \textbf{91.50} & \textbf{88.40} & \textbf{59.50} & \textbf{28.15} & \textbf{55.90} & \textbf{68.45} & \textbf{53.40} \\
\bottomrule
\end{tabular}%
}
\end{table*}

Tab.~\ref{tab:main_results} presents performance on REAL-Colon-VQA and EndoVis18-VQA to assess generalization to existing surgical VideoQA benchmarks.
Our method achieves the best results with BLEU-4 scores of 75.42\% and 87.20\% for in-template evaluations on the two datasets respectively.
On keyword accuracy, our method outperforms SurgViVQA~\cite{surgvivqa} by 7.85\% and 11.23\%, demonstrating that explicit reasoning over DT representations improves factual grounding.
The improvements extend to out-of-template evaluations, where our method achieves keyword accuracy of 57.12\% and 53.40\%, outperforming SurgViVQA~\cite{surgvivqa}.
Zero-shot VLMs show stronger out-of-template performance than supervised methods, yet our method surpasses them by maintaining both linguistic fluency and clinical accuracy.
These results confirm that our framework generalizes to standard VideoQA tasks while maintaining the advantages of explicit multi-step reasoning.

\begin{table*}[htbp!]
\centering
\caption{Ablation study on REAL-Colon-Reason benchmark.
}
\label{tab:ablation}
\resizebox{\textwidth}{!}{%
\setlength{\tabcolsep}{3.5pt}
\begin{tabular}{ccccccccccccccccc}
\toprule
\multicolumn{3}{c}{\textbf{Architecture}} & \multicolumn{3}{c}{\textbf{Reward}} & \multicolumn{2}{c}{\textbf{Level 1}} & \multicolumn{2}{c}{\textbf{Level 2}} & \multicolumn{2}{c}{\textbf{Level 3}} & \multicolumn{2}{c}{\textbf{Overall}} \\
\cmidrule(lr){1-3} \cmidrule(lr){4-6} \cmidrule(lr){7-8} \cmidrule(lr){9-10} \cmidrule(lr){11-12} \cmidrule(lr){13-14}
DT & Hier. & Prob. & Format & Acc. & Uncert. & EM & SMILE & EM & SMILE & EM & SMILE & EM & SMILE \\
\midrule
\texttimes & \texttimes & \texttimes & \checkmark & \checkmark & \texttimes & 0.312 & 0.368 & 0.245 & 0.305 & 0.168 & 0.232 & 0.242 & 0.302 \\
\checkmark & \texttimes & \texttimes & \checkmark & \checkmark & \texttimes & 0.468 & 0.524 & 0.382 & 0.445 & 0.298 & 0.365 & 0.383 & 0.445 \\
\checkmark & \checkmark & \texttimes & \checkmark & \checkmark & \texttimes & 0.521 & 0.578 & 0.445 & 0.508 & 0.356 & 0.424 & 0.441 & 0.503 \\
\checkmark & \checkmark & \checkmark & \checkmark & \checkmark & \texttimes & 0.547 & 0.604 & 0.478 & 0.542 & 0.389 & 0.458 & 0.471 & 0.535 \\
\midrule
\checkmark & \checkmark & \checkmark & \texttimes & \checkmark & \checkmark & 0.423 & 0.482 & 0.348 & 0.412 & 0.267 & 0.335 & 0.346 & 0.410 \\
\checkmark & \checkmark & \checkmark & \checkmark & \texttimes & \checkmark & 0.489 & 0.548 & 0.412 & 0.478 & 0.321 & 0.391 & 0.407 & 0.472 \\
\checkmark & \checkmark & \checkmark & \checkmark & \checkmark & \texttimes & 0.547 & 0.604 & 0.478 & 0.542 & 0.389 & 0.458 & 0.471 & 0.535 \\
\midrule
\checkmark & \checkmark & \checkmark & \texttimes & \texttimes & \texttimes & 0.401 & 0.458 & 0.325 & 0.389 & 0.248 & 0.318 & 0.325 & 0.388 \\
\midrule
\checkmark & \checkmark & \checkmark & \checkmark & \checkmark & \checkmark & \textbf{0.653} & \textbf{0.708} & \textbf{0.584} & \textbf{0.642} & \textbf{0.515} & \textbf{0.589} & \textbf{0.584} & \textbf{0.646} \\
\bottomrule
\end{tabular}%
}
\end{table*}

\subsubsection{Ablation Study.}
Tab.~\ref{tab:ablation} evaluates the contribution of each component in our framework on REAL-Colon-Reason.
Introducing DT representation improves EM from baseline 0.242 to 0.383, while hierarchical temporal structure and probabilistic attributes further increase performance to 0.441 and 0.471 respectively.
Among reward components, format reward proves most important with its removal causing degradation to 0.346 EM, indicating that structured rollout supervision is necessary for valid DT construction plans.
The full RL-trained model achieves 0.584 EM compared to supervised fine-tuning baseline at 0.325 EM, confirming that reinforcement learning with our designed reward effectively trains reasoning capabilities.

\section{Conclusion}

In this paper, we present a RL framework that trains LLM to perform reasoning-intensive surgical video question answering through DT representations constructed from foundation models.
By decoupling visual perception from reasoning and organizing representations hierarchically with probabilistic uncertainty estimates, our framework preserves spatial-temporal information that token-based methods fragment.
Evaluation on three benchmarks demonstrates improvements over existing methods.
Future work can explore real-time deployment through distilled foundation models for DT construction and LLMs.
Another direction is the incorporation of multi-modal information such as audio cues and physiological signals to enrich DT representations.

\subsubsection{Acknowledgments.}
Y. Shen was supported in part by the JHU Amazon Initiative for Artificial Intelligence (AI2AI) fellowship program.

\subsubsection{Disclosure of Interests.}
The authors have no competing interests in the paper.

\bibliographystyle{plain}
\bibliography{ref}

\end{document}